\documentclass[10pt,twocolumn,letterpaper]{article}

\usepackage{cvpr}
\usepackage{times}
\usepackage{epsfig}
\usepackage{graphicx}
\usepackage{amsmath}
\usepackage{amssymb}
\usepackage{multirow}
\usepackage{threeparttable}
\usepackage{array}
\usepackage{footmisc}

\usepackage[breaklinks=true,bookmarks=false]{hyperref}

 \cvprfinalcopy 

\def\cvprPaperID{5809} 

\begin{document}

\title{Texture Deformation Based Generative Adversarial Networks for Face Editing}

\author{WenTing Chen \footnote{Equal contribution}\\
Institute of Computer Vision\\
Shenzhen University\\
{\tt\small chenwenting2017@email.szu.edu.cn}
\and
Xinpeng Xie \footnotemark[1]\\
Institute of Computer Vision\\
Shenzhen University\\
{\tt\small xiexinpeng2017@email.szu.edu.cn}
\and
Xi Jia\\
Institute of Computer Vision\\
Shenzhen University\\
{\tt\small jiaxi@email.szu.edu.cn}
\and
Linlin Shen \footnote{Corresponding author}\\
Institute of Computer Vision\\
Shenzhen University\\
{\tt\small llshen@szu.edu.cn}
}
\date{}
\maketitle

\begin{figure*}[h]
\begin{center}
\includegraphics[width=1.0\linewidth]{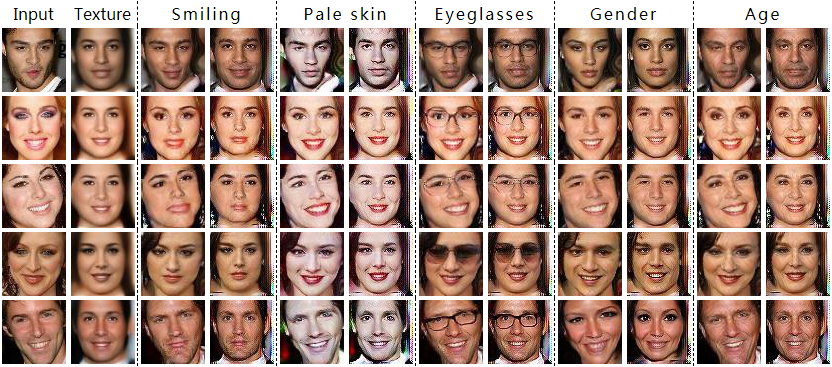}
\end{center}
   \caption{TDB-GAN transfer attributes by texture based deformation. The first and second columns are the input images and the texture generated by DAE, respectively. In the remaining columns, while the even columns show the textures produced by generator for the target attribute, the odd ones show the final synthesized images by wrapping the texture.}
\label{fig:img_texture_celeba}
\end{figure*}

\begin{abstract}

 Despite the significant success in image-to-image translation and latent representation based facial attribute editing and expression synthesis, the existing approaches still have limitations in the sharpness of details, distinct image translation and identity preservation. To address these issues, we propose a Texture Deformation Based GAN, namely TDB-GAN, to disentangle texture from original image and transfers domains based on the extracted texture. The approach utilizes the texture to transfer facial attributes and expressions without the consideration of the object pose. This leads to shaper details and more distinct visual effect of the synthesized faces. In addition, it brings the faster convergence during training. The effectiveness of the proposed method is validated through extensive ablation studies.  We also evaluate our approach qualitatively and quantitatively on facial attribute and facial expression synthesis. The results on both the CelebA and RaFD datasets suggest that Texture Deformation Based GAN achieves better performance.
\end{abstract}

\section{Introduction}

Face editing aims to change or enhance facial attributes (e.g. hair color, expression, gender and age), and add virtual makeup to human faces. In recent years, face editing has attracted great interests in computer vision fields \cite{Chen2018Facelet,Xiao_2018_ECCV,shu2017neural}. Several image-to-image translation methods \cite{isola2017image,CycleGAN2017,yi2017dualgan} have achieved facial attributes and expressions manipulation on single or multiple domains. Most methods are based on the generative adversarial networks (GANs) \cite{goodfellow2014generative} like Cycle GAN \cite{CycleGAN2017}, IcGAN \cite{Perarnau2016}, StarGAN \cite{StarGAN2018}, etc.  However, the generators of most image-to-image translation approaches are fed with the input image directly. While altering the facial expressions, synthesized expressions are not genuine enough since they only modify the face moderately.

Moreover, the task of face editing can also be tackled with the Encoder-Decoder architecture through decoding the latent representation from encoder conditioned on target attributes. This kind of architecture aims to figure out the relationship between the facial attributes and the latent representation and impose the latent representation to face editing \cite{DBLP:conf/bmvc/ZhouXYFHH17,xiao2018dna,sun2018mask,natsume2018rsgan}. Commonly, most Encoder-Decoder architectures encode the input image to a low-dimension latent representation which may lead to the loss of information and representation capability. Besides, most approaches fail to preserve the identity during face editing.
	
To solve the issues raised from both image-to-image translation and latent representation approaches, we first adopt the DAE model \cite{shu2018deforming} to transfer the input image to three different physical image signals, including shading, albedo and deformation with an Encoder-Decoder architecture, and then combine shading and albedo to generate a pure and well-aligned texture image that presents the illumination effects and the characteristic appearance of the face. Then, we feed both the generated texture and target domain labels into a GAN model to synthesize a new texture image with target attributes. Finally, we warp the generated texture with the spatial deformation to generate the ultimate result, we also employ an identity loss between the generated image and input image to preserve identity. Overall, our main contributes are summarized as follows:

1. We propose the Texture Deformation Based GAN, a novel framework that learns the mappings among multiple domains based on disentangled texture and warps the generated texture spatially to generate the face image with target domain features.

2. We empirically demonstrate the effectiveness of our TDB-GAN through the ablation studies on facial attribute editing and expression synthesis. We validate the superiority of texture-to-image translation over the image-to-image translation. We also prove the effectiveness of identity loss through the face verification.

3. The proposed TDB-GAN is evaluated on facial attributes and expression synthesis both qualitatively and quantitatively. The results suggest TDB-GAN outperforms the existing methods.

\section{Related works}

The popularity of generative models has a great effect on face editing. The Encoder-Decoder architecture and Generative Adversarial Network (GAN) \cite{goodfellow2014generative} are the two major categories of methods for this task.

\textbf{Intrinsic Deforming Autoencoder (DAE)} \cite{shu2018deforming} is a novel generative model which decomposes the input image into texture and deformation. DAE follows the deformable template paradigm and models image generation through texture synthesis and spatial deformation. DAE can obtain the prototypical object by removing the deformation. Discarding variability due to deformations, the texture encoded from the original image is a purer representation. Moreover, by modeling the face image in terms of a low-dimensional latent code, we can more easily control the facial attributes and expression over the generative process.

\textbf{Generative Adversarial Networks (GANs)}  \cite{goodfellow2014generative} is a promising generative model and can be used to solve various computer vision tasks such as image generation  \cite{huang2017stacked,yan2016attribute2image,xian2017texturegan}, image translation  \cite{isola2017image,CycleGAN2017,yi2017dualgan}, and face image editing \cite{Xiao_2018_ECCV,StarGAN2018,natsume2018rsgan}. The GAN model is mainly designed to learn a generator G to generate fake samples and a discriminator D to distinguish between real and fake samples. Besides leveraging the typical adversarial loss, a reconstruction loss is often employed \cite{StarGAN2018,he2017arbitrary} to generate the faces as realistic as possible. Additionally, an identity loss is proposed to assure that the generated faces preserve the original identity in our approach.

\textbf{Pix2Pix} \cite{isola2017image} is a typical image-to-image translation based method. The approach can learn the mapping between input and output domains and has achieved impressive results in several image translation tasks \cite{CycleGAN2017,yi2017dualgan,liu2017unsupervised}.   Pix2Pix combines adversarial loss with L1 loss to transfer images in a paired way. For unpaired images, several frameworks like MUNIT  \cite{huang2018munit}, CycleGAN \cite{CycleGAN2017}, and Invertible Conditional GAN \cite{Perarnau2016} have been proposed. However, all the frameworks try to learn the joint distribution between two domains, which limits them to handle multiple domains at the same time.

\textbf{StarGAN} \cite{StarGAN2018} is the first generative model to achieve multi-domain image-to-image translation across different datasets with only one single generator. It also consists of two modules, a discriminator D to distinguish between real and fake images and classify the real images to its corresponding domain, and a generator G generates a fake image using both the image and target domain label (binary or one-hot vector). One of the novelties in StarGAN is that its generator G is allowed to reconstruct the original image from the fake image given the original domain label. StarGAN also utilizes a mask vector with the domain label to enable joint training between domains of different datasets. However, StarGAN is an image-to-image model and does not involve any latent representation, so its capability of changing facial attributes is limited.

\textbf{AttGAN} \cite{he2017arbitrary} is a multiple facial attribute editing model that contains three components at training: the attribute classification constraint, the reconstruction learning and the adversarial learning. The content that latent representation deliveries is uncertain and limited. Hence, imposing the attribute label to the latent representation might change other unexpected parts. Similar to StarGAN, AttGAN applies an attributes classification constraint to guarantee the correct attribute manipulation on the generated image and a reconstruction learning to preserve the attribute-excluding details. AttGAN tries to free the attribute-independent constraint from the latent representation, while our approach encodes the input to different latent representation to generate texture and employ an image-to-image translation to achieve face editing.

\begin{figure*}[t]
\begin{center}
\includegraphics[width=1.0\linewidth]{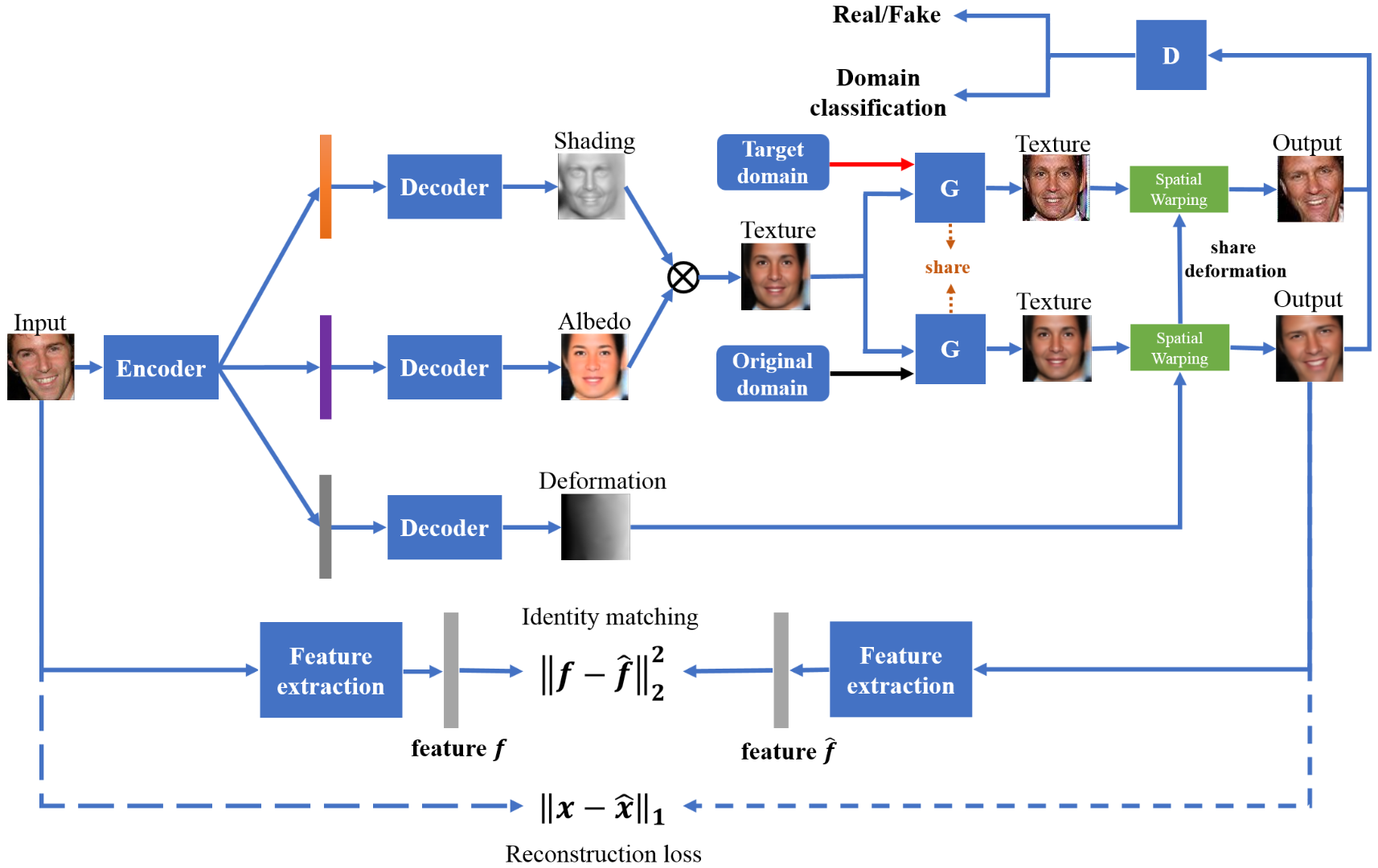}
\end{center}
   \caption{Overview of Texture Deformation Based GAN.}
\label{fig:overview}
\end{figure*}

\section{Texture Deformation Based GAN}

In this section, we introduce the Texture Deformation Based GAN (TDB-GAN) framework for face attributes editing. As shown in Figure \ref{fig:overview}, the TDB-GAN consists of two major modules, i.e. the intrinsic deforming autoencoder DAE and GAN based image-to-image translation module.
\subsection{Intrinsic Deforming Autoencoder}
The recent works \cite{StarGAN2018} aim at translating an original face image to a new face image with different attributes. However, the pose and shape of face might have influence on facial attributes and facial expression synthesis. Thus, we utilize the Intrinsic DAE \cite{shu2018deforming} to separate a face image into texture and deformation to disentangle the variation. DAE adopts the intrinsic decomposition regularization loss to model the physical properties of shading and albedo. The shading and the albedo are then combined to generate the texture that eliminates the geometric information and can represent the identity, illumination, face attributes and so on, whereas the deformation describes the spatial gradient of the warping field (spatial transformation).

\subsubsection{The architecture of encoder}
In this module, we feed the encoder $E_{\theta_{enc}}$ , a densely connected convolutional network, with an input image $I_{Input}$. Then, it generates a latent representation Z for the following decoders. Particularly, the latent representation can be decomposed as follows: Z=[$Z_S$,$Z_A$,$Z_D$ ], where $Z_S$, $Z_A$ and $Z_D$ are shading-related, albedo-related and deformation-related representations, respectively.
\begin{equation}
Z=E_{\theta_{enc}}(I_{Input})
\end{equation}

\subsubsection{Decomposition of shading, albedo and deformation}

As visualized in Figure \ref{fig:overview}, we introduce three separate decoders for shading, albedo and deformation, including $D_S$, $D_A$ and $D_D$. The inputs to these decoders are delivered by a joint encoder network. The shading-related, albedo-related and deformation-related decoders are fed with the latent representations $Z_S$, $Z_A$ and $Z_D$ respectively. The decoders can provide us with a clear separation of shading, albedo and deformation. The equation can be written as follows:
\begin{equation}
S= D_S \left(Z_S \right)
\end{equation}
\begin{equation}
A= D_A \left(Z_A \right)
\end{equation}
\begin{equation}
De= D_D\left (Z_D \right)
\end{equation}
where $S$, $A$ and $De$ denote shading, albedo and deformation.
Then the texture $T$ of the input image can be computed by the shading $S$ and albedo $A$ with the Hadamard product:

\begin{equation}
T=S \odot A
\end{equation}

Finally, the generated texture is warped spatially with the deformation to synthesize the ultimate image $I_{Output}$. W denotes the operation of spatial warping.
\begin{equation}
			I_{Output}=W\left(T,De \right)
\end{equation}

\subsubsection{The objective function}

The objective function is composed of three losses, including $L_R$, $L_{smooth}$ and $L_B$. It can be written as:
\begin{equation}
L_{DAE}=L_R+L_{smooth}+L_B+L_{Shading},
\end{equation}
where the reconstruction loss is defined as:
\begin{equation}
L_R={\left \| I_{Ouput}-I_{Input}  \right \|}_2 ,
\end{equation}
the smoothness cost is given by:
\begin{equation}
L_{smooth}=\lambda_1 (\left \| \nabla W_x (x,y)\right \|_1+\left \| \nabla W_y (x,y)\right \|_1 ),
\end{equation}
the bias reduce loss is formatted as:
\begin{equation}
L_B=\lambda_2 \left \| S_A-S_0 \right \| ^2+{\lambda_2}^\prime \left \|\bar{W}-W_0 \right \| ^2
\end{equation}
and the shading loss is written as:
\begin{equation}
L_{Shading}=\lambda_3 \left \| \nabla S \right \|^2
\end{equation}

In the equations above,  $I_{Input}$ and $I_{Ouput}$ represent the input images and reconstructed images respectively. $ \nabla W_x (x,y)$  and $ \nabla W_y (x,y)$ stands for local warping field. $S_0$ and $S_A$ denote the identity and average affine transform within the minibatch. $W_0$ and $\bar{W}$ represent identity grid and average deformation grid within a minibatch.

\subsection{Multi-Domain Texture-to-Image Translation}

Similar to StarGAN, our goal is to train a multi-domain texture-to-image translation network. We first feed the generator with texture and target domain  labels randomly sampled from training data. Then, we warp the generated texture with deformation to synthesize the fake face image. We also impose the domain classification loss to classify the domain of the fake face image. Furthermore, we use the reconstruction loss and identity loss to supervise the generator to synthesize more realistic and identity-preserved face images, respectively.

\subsubsection{Adversarial loss}

We utilize the adversarial loss to enable the generated images as genuine as the real samples. The adversarial loss can be written as:

\begin{equation}
\begin{split}
L_{adv}=E_x \left[logD_{src} \left(x \right) \right ] \\ +E_{t,c} \left [log \left(1-D_{src} \left(W(G\left(t,c\right),De\right)\right) \right]
\end{split}
\label{eq:adv}
\end{equation}
In this loss function, $G$ generates a new texture $G(t,c)$ conditioned on both the face texture t and target domain label $c$, while $D$ strives to differentiate the real face texture from the generated face texture. In Eq.$ \left ( \ref{eq:adv} \right )$, $D_{src}(x)$ denotes a probability distribution over sources given by $D$. The discriminator tries to maximize this objective, whereas the generator tries to minimize it.

\subsubsection{Domain classification loss}
To enable the generator to generate the fake image with the target domain, we add a domain classifier on the top of $D$. For the optimization of $D$ and $G$, we define the domain classification of the real image as follow:
\begin{equation}
L_{cls}^r = E_{t,c^\prime}  \left [-logD_{cls}  \left(c^\prime |x  \right)\right],
\end{equation}
where $c^\prime$ stands for the original domain label for the real face image. The term $D_{cls} (c^\prime |x)$ represents a probability distribution over domain labels produced by $D$. In addition, the domain classification loss of the fake face texture is defined as
\begin{equation}
L_{cls}^f=E_{t,c} \left[-logD_{cls} \left (c|W \left (G \left(t,c \right),De \right) \right) \right] .
\end{equation}

\subsubsection{Reconstruction loss}

By optimizing the adversarial and classification loss, $G$ is able to generate the realistic face texture with proper attributes. Nonetheless, we cannot guarantee that the generated face texture preserves the content of the input face texture while changing the domain-related parts of the input face texture. Therefore, the reconstruction loss is imposed to the reconstructed texture and image, respectively. For the texture image, we apply a cycle consistency loss proposed by Zhu et al. \cite{CycleGAN2017} to our generator, which is defined as:
\begin{equation}
L_{rec}^t=E_{t,c,c^\prime} \left [ \left \|t-G \left (G \left (t,c  \right ),c^\prime \right )\right \|_1 \right ],
\end{equation}
where $G$ takes the generated face texture $G(t,c)$ and the original domain label $c^\prime$ as input and tries to reconstruct the original face texture. We utilize the L1 norm to compute our reconstruction loss.

For the reconstructed image, the generator synthesizes the new texture with the original texture t and domain label  $c^\prime$. Then, the new texture is warped with deformation to generate the output image.  L1 norm of the difference between the input and the generated image is defined as below:
\begin{equation}
L_{rec}^i=E_{t,c,c^\prime} \left [\left \| x-W \left (G \left (t,c^\prime  \right ),De \right )\right \|_1 \right ].
\end{equation}
\begin{equation}
L_{rec}=L_{rec}^t+L_{rec}^i
\end{equation}

\subsubsection{Identity loss}
Even though reconstruction loss can preserve some unrelated content of the input face texture, the generator might still change the identity of the output face texture. The generator would not only learn the attribute relative parts but also learns the identity corresponding to the person with label c from the training set. For example, majority of celebrities of face images in CelebA \cite{liu2015deep} come from Europe or America, and only few are from Asia. Therefore, when learning the attributes from European or American, Asian might not preserve its own particular facial features.

Therefore, we exploit an identity preserving network $F_{ip}$ to retain the identity discrimination of the synthesized face texture, and an identity loss $L_{ip}$ to preserve personal facial features. This approach is derived from the work proposed by Huang  \cite{huang2017beyond}. $F_{ip}$ denotes a feature extractor to extract the feature of the synthesized face texture $\hat{t}$ and the real face texture $t$. We select the LightCNN  \cite{wu2018light} as our feature extractor and fix the parameters in the training procedure. Specifically, we apply the output of the second to last fully connected layer of $F_{ip}$ to the identity loss $L_{ip}$:
\begin{equation}
L_{ip}= \left \|  F_{ip}(t) - F_{ip}(\hat{t}) \right \|_2 ^2
\end{equation}
where $ \left \| \cdot \right \|_2 $ denotes the L2-norm.

\subsubsection{GAN-related objective function}
Overall, the final objective functions to optimize $G$ and $D$ are illustrated as:
\begin{equation}
L_D=-L_{adv}+\lambda_{cls} L_{cls}^r
\end{equation}
\begin{equation}
L_G=L_{adv}+\lambda_{cls} L_{cls}^f+\lambda_{rec} L_{rec}+\lambda_{ip} L_{ip},
\end{equation}
where $\lambda_{cls}$, $\lambda_{rec}$ and $\lambda_{ip}$ are hyper-parameters to control the weight of domain classification, reconstruction and identity loss.

\section{Implementation}
In this section, we demonstrate how we stabilize the training process and the details of the network architecture.

\subsection{Network Architecture}
Since the proposed TDB-GAN consists of two major modules, we directly utilize the encoder and decoder architectures from DAE \cite{shu2018deforming}. The generator and discriminator architectures are adopted from StarGAN \cite{StarGAN2018} in our framework. We also leverage PatchGANs\cite{isola2017image,CycleGAN2017} for the discriminator to distinguish the real images from synthesized images.
\subsection{Training Strategy}
In order to stabilize and accelerate the training procedure of TDB-GAN, we propose a multi-stage training strategy. In the first stage, we only optimize the DAE model, namely the $L_R$, $L_{smooth}$, $L_B$ and $L_{Shading}$. Then, we fix the pretrained weights of DAE model. Simultaneously, the generator G and discriminator D are trained with the $L_G$ (with $\lambda_{ip}=0$) and $L_D$ loss,  respectively. Finally, we jointly train $L_{DAE}$, $L_G$ and $L_D$. Note that, we impose the identity loss $L_{ip}$ in the final training stage to ensure that the generated image preserves the identity.

\section{Experiments}
In this section, we first compare TDB-GAN with and without the DAE module on facial attribute transfer. In addition, we demonstrate empirical results that the result of TDB-GAN with identity loss can preserve more identity information than that without it.

\subsection{Datasets}

\textbf{The CelebFaces Attributes (CelebA) dataset} \cite{liu2015deep} contains 202,599 face images of 10,177 celebrities, each annotated with 40 binary attributes. We resize all aligned images from the  $178\times 218$   into $64	\times 64$. We randomly select 2,000 images as test set and use the remaining images for training. We mainly test ten domains with the following attributes: expression (smiling/not smiling), skin color (pale skin/normal skin), accessory (eyeglasses/no eyeglasses), gender (male/female) and age (young/old).

\textbf{The Radboud Faces Database (RaFD)} \cite{langner2010presentation} consists of 4,824 images collected from 67 subjects. Each subject has eight facial expressions in three different gaze directions, which are captured from three different angles. We first detect all face images with MTCNN \cite{zhang2016joint} and crop out the images with size $384 \times 384 $ resolution, where the faces are centered, and resized to $64 \times 64$.

\subsection{Training}
All the models are optimized with Adam \cite{kingma2014adam}, where $\beta_1 =0.5$ and $\beta_2=0.999$. We flip the images horizontally with a probability of 0.5 to augment the training set. We perform one generator update after five discriminator updates as described in \cite{StarGAN2018}. The batch size is set to 100 for all experiments. For the experiments on the CelebA, we first train the DAE module for 5 epochs with a learning rate of 0.0002. Then, we train the generator and discriminator with a learning rate of 0.0001 for the first 100 epochs and linearly decay the learning rate to 0 over the next 100 epochs. Next, we impose the identity loss to the GAN module and train the GAN-related part for 29 epochs with a learning rate of 0.0001 and apply the aforementioned decaying strategy over the next 29 epochs. The train strategy of RaFD is similar to that of CelebA. The weight in training objective is set as $\lambda_{cls}=1$ for $L_{cls}^r$, $L_{cls}^f$, $\lambda_{rec}=10$ for $L_{rec}$, $\lambda_{ip}=0.001$ for $L_{ip}$, $\lambda_1=1e-6$ for $L_{smooth}$, $\lambda_2=\lambda_2^\prime =0.01$ for $L_B$ and $\lambda_3=1e-6$ for $L_{Shading}$.

\subsection{Ablation studies}
A unique advantage of the TDB-GAN is its capability of disentangling the texture from the input image and editing the facial attributes and expression without impact of the pose and shape of face We conduct an experiment on TDB-GAN with/without DAE module. Additionally, we prove that the proposed identity loss helps to preserve more identity information through a verification result.

\subsubsection{Results with/without DAE}
In TDB-GAN, we prefer to separate the texture and deformation from the input image. We propose that the information of deformation would significantly affect the quality of face editing and the convergence of the domain classification loss of fake face texture during training.
\begin{figure}[t]
\begin{center}
\includegraphics[width=0.9\linewidth]{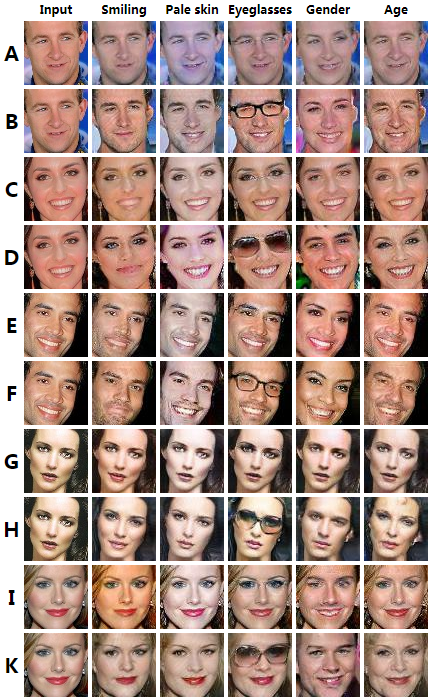}
   \caption{Facial attribute transfer results on the CelebA dataset. The first column demonstrates the input image, next five columns show the single attribute transfer results.  The odd rows display the results generated by the TDB-GAN without DAE module, while the even rows show the results produced with DAE.}
\label{fig:withoutDAE}
\end{center}
\end{figure}
As illustrated in Figure \ref{fig:withoutDAE}, the eyeglasses generated by TDB-GAN with DAE are more obvious. For example, no glasses can be observed for the faces in row C, E, G and I generated by TDB-GAN without DAE. The images generated by TDB-GAN without DAE (A, C) do not show the pale skin as realistic as those generated by TDB-GAN. While the faces of C and E generated by TDB-GAN without DAE are still smiling, TDB-GAN with DAE transfers the face image to smile or not smile correctly and naturally. Lastly, our proposed method has more genuine changing of feminization, masculinity, aging and rejuvenation than the TDB-GAN without DAE module. The main reason is that DAE disentangles texture and deformation from the input image. The former preserves the main feature and identity of the face, whereas the latter contains the information about the pose of head, the shape of face and so on. While we feed the generator with the well-aligned texture, face editing does not need to consider shape invariance. By contrast, the TDB-GAN without DAE transfers the attributes with the constraint of the shape invariance.
It can be also observed from  Figure \ref{fig:withoutDAE} that TDB-GAN with DAE achieves a lower domain classification loss of fake face textures than the TDB-GAN without DAE. There is a clear margin between the curves in the chart. The lower domain classification loss of fake face textures indicates the better attributes transferring.
\begin{figure}[t]
\begin{center}
\includegraphics[width=0.8\linewidth]{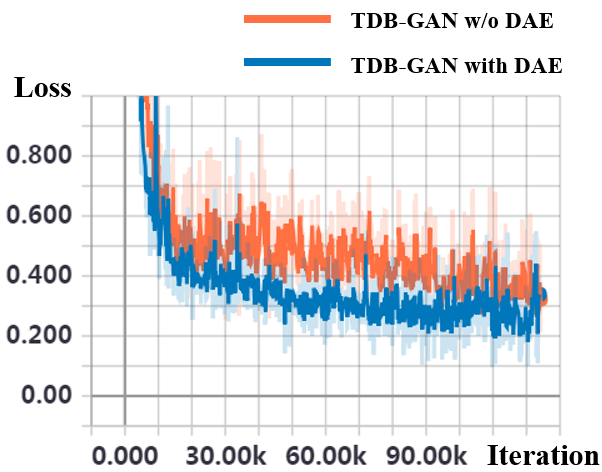}
\end{center}
   \caption{The domain classification loss of the fake face textures generated by the TDB-GAN with/without DAE module.}
\label{fig:losses}
\end{figure}
\subsubsection{Results for identity loss}
While transferring the domains, the network would strive to transfer more average features of the domain to decrease the domain classification loss of the fake face textures even though the domain classification loss of the real face textures plays against it. Thus, we propose the identity loss to ensure the identity invariance. We evaluate the performance of the domain transferring in terms of face recognition accuracy generated by the identity loss. In the following sections, we present the verification results about the TDB-GAN with and without identity loss.

In this experiment, we train our model on RaFD to synthesize facial expressions. There are eight different expressions on RaFD. We fix the input domain as the ‘neutral’ expression and set the target domain to the seven remaining expressions. Thus, the proposed task aims to impose a particular expression to a neutral face.

We randomly split the RaFD dataset into training and testing sets with a 90\%:10\% ratio, namely 4,320 training images and 504 testing images including 63 neutral faces. For each of the neutral face, we apply our network to generate seven facial expression images, i.e. in total 441 fake facial expression images were generated. Based on the 441 generated faces and 504 test images, we randomly generate 3,000 client accesses and 3,000 impostor accesses. The network proposed by Wen and Zhang \cite{wen2016discriminative} is employed to extract 512-dimension identity features from the face images. The cosine distance is adopted to measure the similarity of two faces. The similarity was compared with a threshold (e.g. 0.5) to decide whether they are from the same person, or not. In this work, TPR (True Positive Rate), FPR (False Positive Rate), EER (Equal Error Rate), AP (Average Precision) and AUC (Area under curve) are used to evaluate the performance of face verification. The higher scores of these metrics, except EER, the better results.

\begin{figure}[t]
\begin{center}
\includegraphics[width=0.8\linewidth]{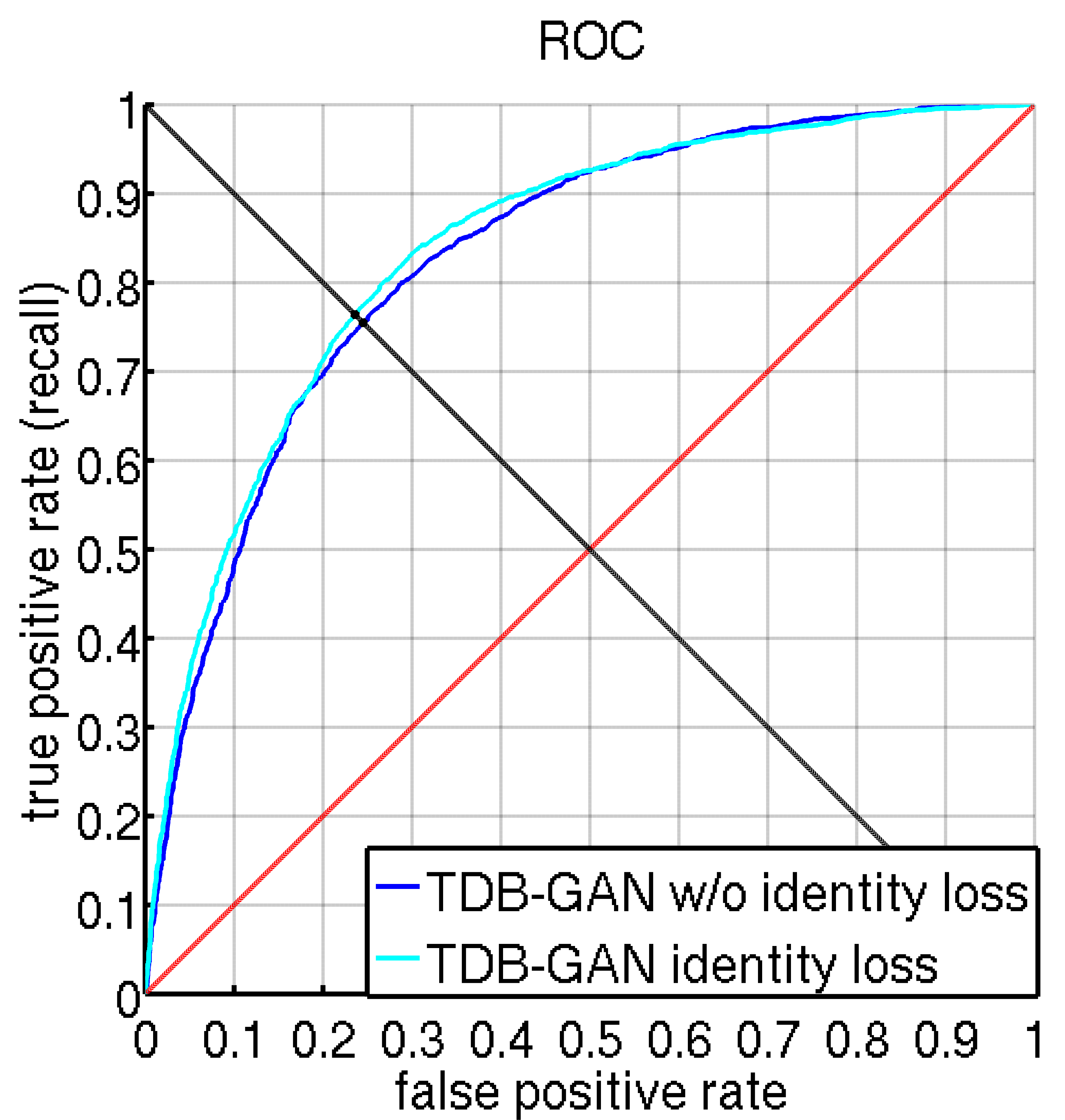}
\end{center}
   \caption{ROC curves on the test set of RaFD dataset.}
\label{fig:ROC}
\end{figure}

Figure \ref{fig:ROC} and Table \ref{table:identityloss} show the ROC curves and the verification results of the TDB-GAN with/without identity loss. From Table \ref{table:identityloss}, while the TPR@FPR=1\% for TDB-GAN without identity loss is 8.70, the identity loss significantly increases the TPR@FRP=1\% to as high as 11.07. Identity loss almost doubles the TPR@FPR=0.1\% and TPR@FPR=0\% of the TDB-GAN. Table \ref{table:identityloss} also suggests that the TDB-GAN with identity loss achieves the lower EER and higher AP and AUC than the TDB-GAN without identity loss.

\begin{table}[t]
\begin{center}
\begin{tabular}{|c|c|c|}
\hline
\multirow{2}{*}{Method } & TDB-GAN  & TDB-GAN\\
 &  with identity loss   &  w/o identity loss  \\
\cline{2-3}
\hline
TPR@FPR=1\% & 11.07 & 8.70   \\
\hline
TPR@FPR=0.1\% & 1.60 & 0.60  \\
\hline
TPR@FPR=0\% & 0.23 & 0.13    \\
\hline
EER (\%) & 23.60 & 24.50     \\
\hline
AP (\%) & 81.89 & 80.29      \\
\hline
AUC (\%) & 83.73 & 82.82     \\
\hline
\end{tabular}
\end{center}
\caption{Verification performance on RaFD dataset.}
\label{table:identityloss}
\end{table}

\begin{figure*}[h]
\begin{center}
\includegraphics[width=0.7\linewidth]{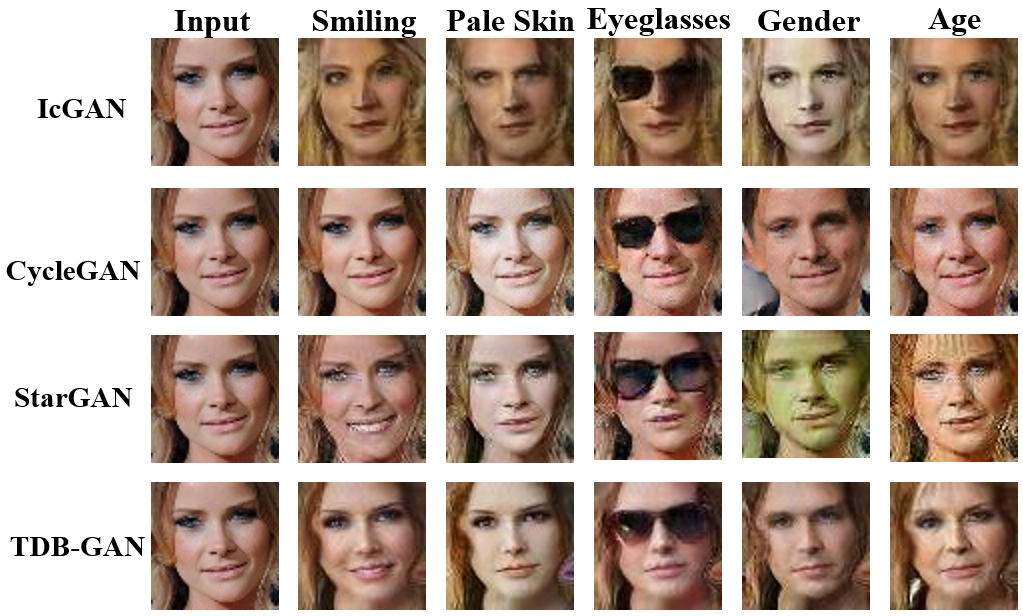}
\end{center}
   \caption{Facial expression synthesis results on CelebA dataset.}
\label{fig:qualityCeleba}
\end{figure*}

\subsection{Qualitative and quantitative evaluation on CelebA}
We first display qualitative results of facial attribute transfer on the CelebA dataset. Then, the quantitative results are evaluated with a user questionnaire.

\subsubsection{Qualitative evaluation}
Figure \ref{fig:qualityCeleba} shows the face images generated by IcGAN, CycleGaN, StarGAN and our TDB-GAN for attribute transfer in smiling, pale, eyeglasses, gender and age. As visualized in the Figure, the images generated by image-to-image translation approaches are better than that generated by IcGAN. Our approach contains more information than the low-dimension latent representation and also preserves the attribute-independent information, like hairstyle. The faces generated by TDB-GAN for gender and age transfer are better than that generated by StarGAN, and the eyeglasses added by TDB-GAN are more natural than that added by CycleGAN. Furthermore, our proposed method not only achieves higher visual quality but also preserves the identity related to the input image due to the effect of identity loss.

\subsubsection{Quantitative evaluation}
For quantitative evaluation, we perform a user study on the visual effect of transferred facial attributes to access IcGAN \cite{Perarnau2016}, CycleGAN \cite{CycleGAN2017}, StarGAN \cite{StarGAN2018} and TDB-GAN. Each of the four approaches were applied to transfer the five facial attributes, i.e. smile, pale skin, eyeglasses, gender and age, of faces from twenty individuals. For each of the five attributes transferred for the 20 subjects, four images synthesized by different models were shown to volunteers and they were asked to select the best one, in terms of the realism, preservation of identity and quality of the facial attribute synthesis. As a number of 15 volunteers participated the questionnaire, a maximum of $20 \times 15 =300$ votes can be received for each approach and attribute. Table \ref{table:quanCelebA} lists the ratio of votes received for each model and attribute. While StarGAN received the highest votes for pale skin transfer, our TDB-GAN received the highest votes for four of the five attributes, i.e. smile, eyeglasses, gender and age.

\begin{table}[h]
\begin{center}
\begin{tabular}{|p{1.45cm}<{\centering}|p{0.85cm}<{\centering}|p{1.4cm}<{\centering}|p{1.25cm}<{\centering}|p{1.33cm}<{\centering}|}
\hline
Models & IcGAN & CycleGAN & StarGAN & TDB-GAN \cr
\hline
Smile & 2.33\% & 21.33\% & 19.00\% & 57.33\%         \cr
\hline
Pale skin & 2.00\% & 37.00\% & 36.67\% & 24.33\%     \cr
\hline
Eyeglasses & 0 &28.00\% & 30.33\% & 41.67\%      \cr
\hline
Gender & 1.33\% & 35.00\% &9.67\% & 54.00\%       \cr
\hline
Age & 0.33\% & 20.00\% & 17.67\% &62.00\%             \cr
\hline
\end{tabular}
\end{center}
\caption{The perceptual evaluation of different models. Note that, the sum of probability of each row is not strictly equal to 100\% due to numerical precision loss.}
\label{table:quanCelebA}
\end{table}
\begin{figure*}[t]
\begin{center}
\includegraphics[width=0.8\linewidth]{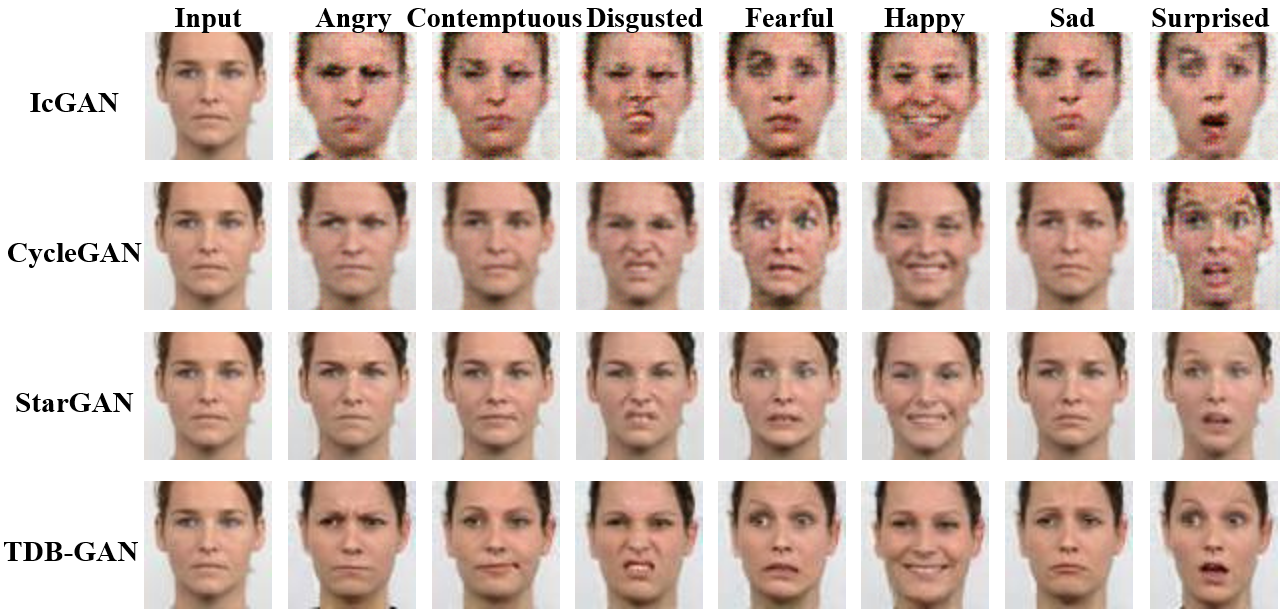}
\end{center}
   \caption{Facial expression synthesis results on RaFD dataset.}
\label{fig:qualityRafd}
\end{figure*}
\subsection{Qualitative and quantitative evaluation on RaFD}
In the following sections, we demonstrate the qualitative and quantitative evaluation results on the RaFD dataset.
\subsubsection{Qualitative evaluation}
Figure \ref{fig:qualityRafd} shows an example of seven facial expressions synthesized by IcGAN \cite{Perarnau2016}, CycleGAN \cite{CycleGAN2017} and StarGAN \cite{StarGAN2018} and our TDB-GAN. As shown in the Figure, the images generated by StarGAN and our TDB-GAN have better visual quality than that generated by IcGAN and CycleGAN. IcGAN transfers the neutral expression to various expressions, but the generated fake images have lowest quality. We believe that the latent vector extracted from IcGAN lacks effective representability. While the performance of CycleGAN is considerably better than that of IcGAN, the fake images generated by CycleGAN are still ambiguous. The fake faces synthesized by StarGAN have much more natural and more distinct expressions. Nonetheless, TDB-GAN is superior to StarGAN for the sharper details and the more distinguishable expressions. For example, the faces generated by our TDB-GAN for angry, fearful and surprised are much more representative than that of StarGAN, especially in the eye regions.

We believe the ability of separating the texture and deformation of TDB-GAN contributes most to the image quality, which allows TDB-GAN to pay more attention to the face expression editing, instead of the pose, shape and so on.

\subsubsection{Quantitative evaluation}
For a quantitative evaluation, we compute the classification error of facial expression recognition on the generated images.

We first train a facial expression classifier with the 4,320 training images. And then we train all the GAN models using the same training set.

For testing, we first use the trained GANs to transfer all the neutral expression of the testing images to seven different expressions. Then we use the aforementioned classifier to classify these synthesized expressions. Table \ref{table:quanRAFD} lists the accuracies of the facial expression classifier on the images synthesized by different GAN models.

As shown in Table \ref{table:quanRAFD}, the images synthesized by TDB-GAN model achieves the highest accuracy, which suggests that it synthesizes the most realistic facial expressions compared with the other methods.

\begin{table}[h]
\begin{center}
\begin{tabular}{|c|c|c|c|c|}
\hline
Models & Accuracy (\%) \\
\hline
IcGAN & 91.61          \\
\hline
CycleGAN & 88.44       \\
\hline
StarGAN & 92.06        \\
\hline
TDB-GAN & 97.28         \\
\hline
\end{tabular}
\end{center}
\caption{The expression classification accuracies of images synthesized by different GAN models.}
\label{table:quanRAFD}
\end{table}
\section{Conclusion}
In this paper, we proposed Texture Deformation Based GAN to perform texture-to-image translation among multiple domains. The proposed TDB-GAN can generate images with higher quality and more relative identity compared to the existing methods, due to the disentangled texture and deformation, and the identity loss.



{\small
\bibliographystyle{ieee}

}


\end{document}